\documentclass[sigconf]{acmart}

\usepackage[utf8]{inputenc} 
\usepackage{subfigure}
\usepackage{enumitem}
\usepackage{xspace}
\usepackage{tikz}

\usepackage{pythonhighlight}

\AtBeginDocument{%
  \providecommand\BibTeX{{%
    \normalfont B\kern-0.5em{\scshape i\kern-0.25em b}\kern-0.8em\TeX}}}

\newcommand{\header}[1]{\vspace{1mm}\noindent\textbf{#1}.}

\newcommand{\headerl}[1]{\vspace{1mm}\noindent\textit{#1}.}

\newcommand{\fairbench}{FairPrep\xspace}
\newcommand{\aif}{AIF360\xspace}
\newcommand{\sklearn}{scikit-learn\xspace}

\newcommand{\eg}{e.g.,\xspace}

\newcommand{\etal}{et al.\xspace}

\newcommand{\adult}{\texttt{adult}\xspace}
\newcommand{\german}{\texttt{germancredit}\xspace}
\newcommand{\propublica}{\texttt{propublica}\xspace}
\newcommand{\ricci}{\texttt{ricci}\xspace}

\definecolor{dgray}{gray}{0.35}

\newcommand*\circled[1]{\tikz[baseline=(char.base)]{
  \node[shape=circle,fill=black,inner sep=1pt] (char) {\textcolor{white}{\texttt{#1}}};}}

\begin{document}

\title{}

\title[FairPrep: Promoting Data to a First-Class Citizen in Studies on Fairness-Enhancing Interventions]{FairPrep: Promoting Data to a First-Class Citizen\\ in Studies on Fairness-Enhancing Interventions}

\author{Sebastian Schelter, Yuxuan He, Jatin Khilnani, Julia Stoyanovich}
\affiliation{%
  \institution{New York University}
}
\email{[sebastian.schelter,yh2857,jatin.khilnani,stoyanovich]@nyu.edu}

\begin{abstract}

The importance of incorporating ethics and legal compliance into machine-assisted decision-making is broadly recognized. Further, several lines of recent work have argued that critical opportunities for improving data quality and representativeness,  controlling for bias, and allowing humans to oversee and impact computational processes are missed if we do not consider the lifecycle stages upstream from model training and deployment. Yet, very little has been done to date to provide system-level support to data scientists who wish to develop and deploy responsible machine learning methods.  We aim to fill this gap and present \fairbench{}, a design and evaluation framework for fairness-enhancing interventions.  

\fairbench{} is based on a developer-centered design, and helps data scientists follow best practices in software engineering and machine learning.  As part of our contribution, we identify shortcomings in existing empirical studies for analyzing fairness-enhancing interventions.  We then show how \fairbench{} can be used to measure the impact of sound best practices, such as hyperparameter tuning and feature scaling. In particular, our results suggest that the high variability of the outcomes of fairness-enhancing interventions observed in previous studies is often an artifact of a lack of hyperparameter tuning.  Further, we show that the choice of a data cleaning method can impact the effectiveness of fairness-enhancing interventions. 
\end{abstract} 

\maketitle


\section{Introduction}
\label{sec:intro}

While the importance of incorporating responsibility --- ethics and legal compliance ---  into machine-assisted decision-making is broadly recognized, much of current research in fairness, accountability, and transparency in machine learning focuses on the last mile of data analysis --- on model training and deployment. Several lines of recent work argue that critical opportunities for improving data quality and representativeness,  controlling for bias, and allowing humans to oversee and influence the process are missed if we do not consider earlier lifecyle stages~\cite{Kirkpatrick:2017:AD:3042068.3022181,LehrOhm2017,DBLP:conf/ssdbm/StoyanovichHAMS17,Holstein2019}.  Yet, very little has been done to date to provide system-level support for data scientists who wish to develop, evaluate, and deploy responsible machine learning methods.  In this paper we aim to fill this gap.

We build on the efforts of Friedler \etal~\cite{DBLP:conf/fat/FriedlerSVCHR19} and Bellamy \etal~\cite{Bellamy2018}, and develop a generalizable framework for evaluating fairness-enhancing interventions called \fairbench.  Our framework currently focuses on data cleaning (including different methods for data imputation), and model selection, tuning and validation (including feature scaling and hyperparameter tuning), and can be extended to accommodate earlier lifecycle stages, such as data integration and curation.  In designing \fairbench we pursued the following goals:   

\begin{itemize}[leftmargin=*]
\item Expose a \emph{developer-centered design} throughout the lifecycle, which allows for low effort customization and composition of the framework's components.  Here, we refer to data scientists and software developers.
\item Follow software engineering and machine learning best practices to reduce the \emph{technical debt} of incorporating fairness-enhancing interventions into an already complex development and evaluation scenario~\cite{Sculley2015,Schelter2018c}.  Figure~\ref{fig:bigpicture} summarizes the  architecture of \fairbench.
\item Surface \emph{discrimination} and \emph{due process} concerns, including but not limited to disparate error rates, failure of a model to fit the data, and failure of a model to generalize~\cite{LehrOhm2017}.  
\end{itemize}

In what follows, we further motivate the need for a comprehensive design and evaluation framework for fairness-enhancing interventions, and explain how \fairbench{} can meet this need. 

\subsection{\fairbench by Example}
\label{sec:intro:example}

Consider Ann, a data scientist at an online retail company who wishes to develop a classifier for deciding which payment options to offer to customers.  Based on her experience, Ann decides to include customer self-reported demographic data together with their purchase histories.  Following her company's best practices, Ann will start by splitting her dataset into training, validation and test sets. Ann will then use pandas, scikit-learn, and the accompanying data transformers to explore the data and implement data preprocessing, model selection, tuning, and validation.  To ensure \emph{proper isolation of held-out test data}, Ann will work with the training and validation datasets, not the test dataset, during these stages.

As the first step of data preprocessing, Ann will compute  value distributions and correlations for the features in her dataset, and \emph{identify missing values}. She will fill these in using a default interpolation method in scikit-learn, replacing missing values with the mean value for that feature.  As another preprocessing step, Ann will perform \emph{feature scaling} for the numerical attributes in her data.  This step, also known as normalization, ensures that all features map to the same value range, which will help certain kinds of classifiers up-stream fit the data correctly.

Finally, following the accepted best practices at her company, Ann implements model selection and tuning.  She will identify several classifiers appropriate for her task, and will then \emph{tune  hyperparameters} of each classifier using $k$-fold cross-validation.  To do so, she will specify a hyperparameter grid for each classifier as appropriate, will train the classifier for each point on the grid, and will then use \emph{her company's standard accuracy metrics} to find a good setting of the hyperparameters on the validation dataset.  As a result of this step, Ann will identify a classifier that shows acceptable accuracy, while also exhibiting sufficiently low variance. 

The reader will observe that no fairness issues were surfaced in Ann's workflow up to this point.  This changes when Ann considers the accuracy of her classifier more closely, and observes a disparity: the accuracy is lower for middle-aged women, and for female customers who did not specify their age as part of their self-reported demographic profile.  Ann goes back to data analysis and observes that the value of the attribute {\tt age} is missing far more frequently for female users than for male users. Further, she compares age distributions by gender, and notices differences starting from the mid-thirties.  Ann hypothesizes age to have been an important classification feature, revisits the data cleaning step, and selects a state-of-the-art \emph{data imputation method} such as Datawig~\cite{Biessmann2018} to fill in age (and other missing values) in customer demographics.

Having adjusted data preprocessing in an attempt to reduce error rate disparities, Ann is now faced with several related challenges:
\begin{itemize}[leftmargin=*]
\item How should the data processing pipeline be extended to incorporate additional fairness-specific evaluation metrics?  Hand-coding evaluation metrics on a case-by-case basis, and determining how these should be traded off with each other, and with existing metrics is both time-consuming and error prone.

\item How can the effects of fairness-enhancing interventions be quantified, and judiciously validated, to allow Ann to make an informed choice about which intervention to pick?  These interventions may range from an improved data cleaning method that helps reduce variance for a demographic group, to a fairness-aware classifier, and they may be incorporated at different pipeline stages --- during data preprocessing, immediately before or after a classifier is invoked, or as part of the classification itself.

\item  How does one continue to follow software engineering and ML best practices when incorporating fairness considerations into these pipelines?  For example, how does Ann ensure appropriate level of isolation of the test set?  How does she go about tuning hyperparameters in light of additional objectives?  How does she make her analysis reproducible, to support more effective debugging, and auditing for correctness and legal compliance?
\end{itemize}

To address these challanges, Ann will turn to existing development and evaluation frameworks, that by Friedler et al.~\cite{DBLP:conf/fat/FriedlerSVCHR19} and IBM's AIF360~\cite{Bellamy2018}.  While these frameworks are certainly a good starting point, they will unfortunately fall short of meeting Ann's needs.\footnote{This conjecture was verified by Ann, who met with us for a drink after her failed attempts.  Ann's real name and bar location are suppressed for anonymity~:)}  The main reason is that these frameworks are designed around a small number of academic datasets and use cases, and do not allow to integrate additional data preprocessing steps that are a crucial part of existing machine learning pipelines, and are not designed to enforce best practices.

\subsection{Contributions and Roadmap}
\label{sec:intro:map}

This paper makes the following contributions:
\begin{itemize}[leftmargin=*]
    \item We discuss shortcomings and lack of best practices in existing empirical studies and software for analyzing fairness-enhancing interventions~(Section~\ref{sec:shortcomings}).
     \item We propose \fairbench, a design and evaluation framework that makes data a first-class citizen in fairness-related studies.  \fairbench implements a modular data lifecycle, allowing to re-use existing implementations of fairness metrics and interventions, and to integrate custom feature transformations and data cleaning operations from real world use cases~(Sections~\ref{sec:design} and~\ref{sec:impl}).  We implement \fairbench{} on top of \sklearn{}~\cite{Pedregosa2011} and \aif{}~\cite{Bellamy2018}~(Section~\ref{sec:impl}). 
  \item We apply \fairbench to illustrate that enforcing best practices of machine learning evaluation, which are easy to get accidentally wrong with existing frameworks, and incorporating data cleaning methods can impact the effectiveness of fairness-enhancing interventions~(Section~\ref{sec:study}). We present results of running \fairbench using some of the same benchmark datasets, classifiers, and fairness-enhancing interventions as Friedler \etal~\cite{DBLP:conf/fat/FriedlerSVCHR19}.
\end{itemize}  
  
We present related work in Section~\ref{sec:related} and conclude in Section~\ref{sec:conc}. 

\section{Shortcomings of Previous Work}
\label{sec:shortcomings}

We inspected the code bases for existing studies~\cite{DBLP:conf/fat/FriedlerSVCHR19}, and evaluation frameworks~\cite{Bellamy2018}, and thereby identified a set of shortcomings that motivated us to design a comprehensive, data-centric evaluation framework. In the following, we detail our findings.

\subsection{Insufficient Isolation of Held-Out Test Data}
\label{sec:shortcomings:isolation}

A major requirement for the offline evaluation of ML algorithms is to simulate the real-world deployment scenario as closely as possible. In the real world, we train our model (and select its hyperparameters) on observed data from the past, and predict for target data later, which we have not yet seen and for which we typically do not know the ground truth. In offline evaluation, we typically evaluate a model on a test set that was randomly sampled from observed historical data. It is crucial that this test set be completely isolated from the process of model selection, which, in turn, is only allowed to use training data (the remaining, disjunct observed historical data).
Importantly, data isolation must also be guaranteed for preprocessing operations such as feature scaling or missing value imputation. If these operations were allowed to look at the test set, this could potentially result in a target leakage. 

Unfortunately, we encountered several violations of the test set isolation requirement in the existing benchmarking framework by Friedler at al.~\cite{DBLP:conf/fat/FriedlerSVCHR19}.  These violations, detailed below, bring into question the reliability of reported study results.  Further, we found that the architecture of the IBM AIF360 toolkit~\cite{Bellamy2018} does not support data isolation best practices for feature transformation.

\header{Hyperparameter selection on the test set} The grid search for hyperparameters\footnote{\url{https://github.com/algofairness/fairness-comparison/blob/4e7341929ba9cc98743773169cd3284f4b0cf4bc/fairness/algorithms/ParamGridSearch.py\#L41}} of fairness-enhancing models and interventions in~\cite{DBLP:conf/fat/FriedlerSVCHR19} computes metrics for all hyperparameter candidates on the test set and returns the candidate that gave the best performance. This strongly violates the isolation requirement, as we would not know the ground truth labels for data to predict on in the real world, and therefore could only use a hyperparameter setting that worked well on some previously observed data. 
An evaluation procedure should maintain an additional validation set,  used to select the best hyperparameters, and only evaluate the prediction quality of the resulting single best hyperparameter candidate on the test set, in order to measure how well the model generalizes to unseen data. 

\header{Lack of data isolation for missing value imputation} A common challenge in real world ML scenarios is to handle examples with missing values. Often, this challenge is addressed by applying different missing value imputation techniques~\cite{Schmitt2015} to complete the data. Again, in order to simulate real world scenarios as closely as possible, we should carefully isolate training data from the held-out test data for missing value imputation. If our missing value imputation model were allowed to access test data (and could thereby compute statistics of this data, which is unseen in practice), it would exhibit the potential for accidental target leakage. Unfortunately, this isolation is also not incorporated into the design of existing studies, which invoke the missing value handling logic before computing the train/test split of the data.\footnote{\url{https://github.com/algofairness/fairness-comparison/blob/4e7341929ba9cc98743773169cd3284f4b0cf4bc/fairness/preprocess.py\#L37}}

\header{Lack of data isolation for feature transformation} Analogously to the previously discussed case of missing value imputation, we also need to reliably isolate training data from the held-out test data during feature transformation. Many feature transformation techniques (such as scalers for numerical variables or embeddings of certain attributes) rely on the computation of aggregate statistics over the data. To simulate real world scenarios, it is crucial to only compute these aggregate statistics (``fit the feature transformers'') on the training data. Computing aggregate statistics before conducting the train/validation/test splits can result in target leakage.

We did not find such cases in the existing studies and frameworks, as their feature transformation mostly consists of format changes and the one-hot encoding of categorical variables --- record-level operations that are indepenent of the data splits. Nevertheless, the design of these frameworks does not support isolated feature computations, as the featurization of the data is applied before data splitting.\footnote{\url{https://github.com/IBM/AIF360/blob/master/aif360/datasets/standard_dataset.py\#L84}} Therefore, a data scientist could accidentally introduce target leakage if she followed the existing software architecture.

\subsection{Lack of Hyperparameter Tuning for Baseline Algorithms} 
\label{sec:shortcomings:tune}

We additionally found that the study by Friedler et al.~\cite{DBLP:conf/fat/FriedlerSVCHR19} did not tune the hyperparameters of the baseline algorithms\footnote{\url{https://github.com/algofairness/fairness-comparison/tree/35fb53f7cc7954668eeee28eac5fb20faf89b3d8/fairness/algorithms/baseline}} for which pre-processing and post-processing interventions are applied, even though they tuned the hyperparameters of the fairness interventions, investigating the resulting fairness / accuracy trade-off. This is problematic because there is in general no guarantee that the learning procedure for a baseline algorithm with the default parameters will converge to a solution that fits training data and generalizes to unseen data.  If such a model were deployed in practice, this failure to fit and to generalize would lead to a due process violation according to Lehr and Ohm (see~\cite{LehrOhm2017} p.710-715).

Friedler et al.~\cite{DBLP:conf/fat/FriedlerSVCHR19} found high variability of the fairness and accuracy outcomes with respect to different train/test splits. While we were unable to reproduce these results directly (see Section~\ref{sec:shortcomings:reproduce}), were were able to observe a similar level of variability in our experiments with default parameter settings.

ML textbooks~\cite{Hastie2005} and research~\cite{Kohavi1995} suggests to use more expensive evaluation techniques such as $k$-fold cross-validation, which have the advantage of quantifying the variability of the estimated prediction error for a given hyperparameter selection (and thus giving a principled method to navigate the bias-variance trade-off).

\subsection{Lack of Feature Scaling}
\label{sec:shortcomings:scale}

We observed that both existing frameworks~\cite{DBLP:conf/fat/FriedlerSVCHR19,Bellamy2018} do not normalise the numeric features of the input data, but keep them on their original scale. While some ML models such as decision trees are insensitive features on different scales, many other algorithms implicitly rely on normalized and/or standardized features. Examples among the objective functions of popular models are the RBF kernel used in support vector machines, as well as L1 and L2 regularizers of linear models.

\subsection{Removal of Records with Missing Values}
\label{sec:shortcomings:missing}

Another point of critique is that the study of Friedler et al.~\cite{DBLP:conf/fat/FriedlerSVCHR19} ignored records with missing values in the data (by removing them before running experiments), which means that the studies' findings do not necessarily generalize to data with quality issues. Yet, real-world decision-making systems still have to make decisions for data with missing values. The framework from~\cite{DBLP:conf/fat/FriedlerSVCHR19} has a handle for a dataset to treat missing data, but this is never implemented as far as we could determine. In the default preprocessing routines, records with missing values are always removed from the data.\footnote{\url{https://github.com/algofairness/fairness-comparison/blob/4e7341929ba9cc98743773169cd3284f4b0cf4bc/fairness/preprocess.py\#L40}} 

Thereby, existing frameworks are unable to investigate the effects of fairness enhancing interventions on records with missing values, which could be especially important for cases where a protected group has a higher likelihood of encountering missing values in their data.  It has been documented that survey data from ethnic minorities may be noisier than data collected from the majority ethnic group~\cite{kappelhof2017}.  We also see evidence of this in the benchmark datasets: in the commonly-used Adult Income dataset~\footnote{\url{https://archive.ics.uci.edu/ml/datasets/adult}}, there is a four times higher chance for the \texttt{native-country} attribute to be missing for non-white than for white persons.

\begin{figure*}[t!]
  \centering
  \includegraphics[width=\textwidth]{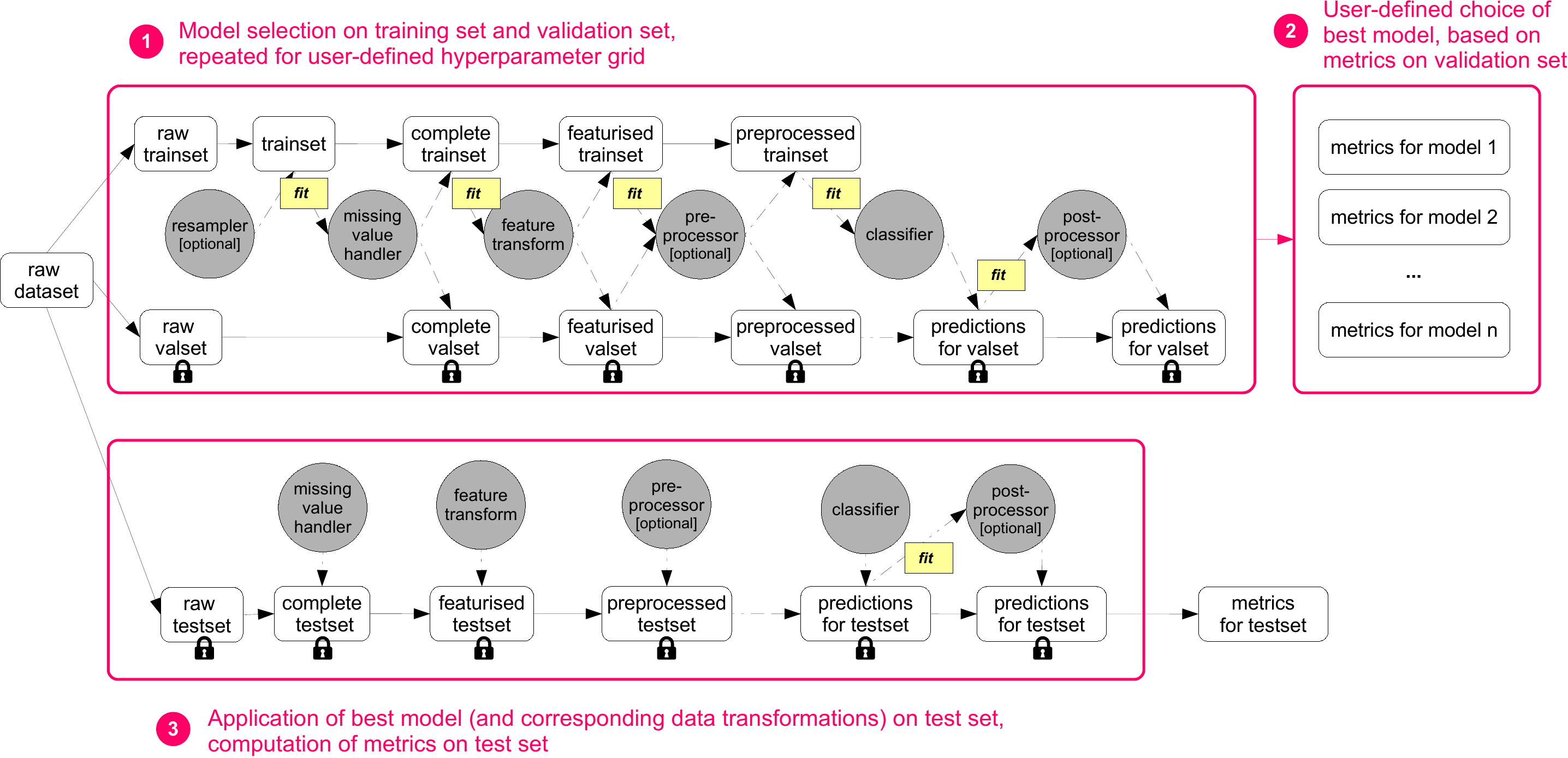}  
  \caption{Data life cycle in \fairbench{}, designed to enforce isolation of the test data, and to allow for customization through user-provided implementations of different components. An evaluation run consists of three different phases: (1)~learn different models, and their corresponding data transformations, on the training set; (2)~compute performance / accuracy-related metrics of the model on the validation set, and allow the user to select the `best' model according to their setup; (3)~compute predictions and metrics for the user-selected best model on the held-out test set.}
  \label{fig:bigpicture}
\end{figure*}

\subsection{Lack of Reproducibility}
\label{sec:shortcomings:reproduce}

An important objective of an evaluation framework should be to make its computations reproducible. A major factor for this is to fix the seeds for pseudo-random number generators throughout the evaluation run, and provide the fixed seed to all components (data splitters, learning algorithms, feature transformations) so that they can leverage it to conduct reproducible random number generation. We find that data splitting\footnote{\url{https://github.com/algofairness/fairness-comparison/blob/35fb53f7cc7954668eeee28eac5fb20faf89b3d8/fairness/data/objects/ProcessedData.py\#L22}} in~\cite{DBLP:conf/fat/FriedlerSVCHR19} does not use fixed random seeds. This has been improved in \aif{}~\cite{Bellamy2018}, where fixed random seeds are used for data splitting. However other components such as the methods to train models\footnote{\url{https://github.com/IBM/AIF360/blob/ca48d6557edf61ddfd112d6199397d9e48ebb6e1/aif360/algorithms/transformer.py}}, do not expose a common random seed. 
\section{Framework Design}
\label{sec:design}

Our goal for \fairbench{} is to provide an evaluation environment that closely mimics real world use cases. 
\begin{enumerate}[label=(\roman*),leftmargin=*]
    \item ~\textit{Data isolation} --- in order to avoid target leakage, user code should only interact with the training set, and never be able to access the held-out test set. User code can train models or fit feature transformers on the training data, which will be applied by the framework to the test set later on. This is a form of \textit{inversion of control}, a common pattern applied in middleware frameworks~\cite{Johnson2009}. The framework should furthermore especially take care of data with quality problems.  For example, it should allow experimenters to isolate the effects of their code on records with missing values by computing metrics and statistics separately for them.
    \item \textit{Componentization} --- different data transformations and learning operations should be implementable as single, exchangable standalone components; the framework should expose simple interfaces to users, allowing them to rapidly customize their experiments with low effort.
    \item \textit{Explicit modeling of the data lifecycle} --- the framework defines an explicit, standardized data lifecycle that applies a sequence of data transformations and model training in a particular, predefined order. Users influence and define the lifecycle by configuring and implementating particular components. At the same time, the framework should also support users as much as possible in applying best practices from machine learning and software engineering.
\end{enumerate}

Figure~\ref{fig:bigpicture} illustrates the data lifecycle during the execution of a run of \fairbench{}, which we now describe in detail. The execution of an evaluation run occurs in the three subsequent phases:

\header{\circled{1} Model selection on training set and validation set}  The purpose of this phase is to train different models (for different hyperparameter settings) on the training data, and compute their corresponding performance metrics on the validation set. \fairbench{} applies a fixed series of consecutive steps in this phase, some of which are optional, and all of which can be customized with dedicated component implementations by our users.
\begin{enumerate}[leftmargin=*]
    \item In the first (optional) step, we allow users to resample the training data: to apply bootstrapping, to balance classes, or to generate additional synthetic examples.
    \item Next, the user has to decide how to treat records with missing values. \fairbench{} offers a set of predefined strategies such as `complete case analysis' (removal of records with missing values) or different imputation algorithms, ranging from simple strategies that fill in the most frequent value of an attribute, to more sophisticated strategies that learn a model tailored to the data for imputation. Note that \fairbench{} enforces that imputation models are learned on the training data only.
    \item After imputation on the raw training data, \fairbench{} applies feature transformations to convert the data into a numeric format suitable for learning algorithms. By default, the framework scales numeric features with a user-chosen strategy, and  one-hot encodes categorical values. If the feature transformers require aggregate statistics from the data, we again ensure that these are only computed on the training dataset. The `fitted' feature transformers are stored in memory afterwards, in order to be applied to the validation set and test set in later phases.
    \item The next (optional) step is the application of a pre-processing intervention to enhance the fairness of the outcome (\eg reweighing the training instances).
    \item Subsequently, \fairbench{} trains a classifier on the training data. This can be a baseline classifier (such as logistic regression) that will be combined with a pre-processing or post-processing fairness-enhancing intervention, or a specialized in-processing model for fairness enhancement.
    \item Next, \fairbench{}  repeats the data transformation conducted so far on the validation set, and applies the trained model to compute predictions for the training and the validation datasets.
    \item In the final (optional) step, users have an opportunity to apply a post-processing intervention to adjust computed predictions in a use-case specific manner.
\end{enumerate}  

\header{\circled{2} User-defined choice of best model} In the second phase, \fairbench{} computes a large set of accuracy and fairness-related metrics for each model based on its predictions for the validation set and training set. A user can then choose the `best' model via a user-defined function, selecting the model with a suitable fairness / accuracy trade-off for their scenario.

\header{\circled{3} Application of the `best' model (and its corresponding data transformations) on test set} In the final phase, \fairbench{} will automatically apply the user-selected best model (and its corresponding data transformations)  on the test set, and provide the user with a final set of metrics. Note that, due to data isolation concerns, the user never gets direct access to the test set.


\section{Implementation}
\label{sec:impl}

In the following, we detail implementation aspects of \fairbench{}. Our framework is based on \aif{}~\cite{Bellamy2018}, from which it leverages the dataset abstraction, metrics and fairness enhancing interventions, as well as on \sklearn{}~\cite{Pedregosa2011}, from which it uses several data transformations and models.

\header{Experiments \& Datasets} At the heart of \fairbench{} is an abstract class for experiments that defines the execution order and lifecycle shown in Figure~\ref{fig:bigpicture}. This class needs to be extended for each experimental dataset. For datasets, we build upon the \texttt{BinaryLabelDataset} abstraction from \aif{}, and make their implementation more flexible by allowing operations like one-hot encoding on different versions by adding feature dimensions for unseen categorical values. This enables \fairbench{} to view data in relational form (as a pandas dataframe) or in matrix form (\eg features as numpy matrix), and to access extensive dataset metadata (\eg sensitive attribute information). 
\fairbench integrates several datasets commonly used in fairness-related studies: 

\adult -- The Adult Income dataset~\footnote{\url{https://archive.ics.uci.edu/ml/datasets/adult}} contains information about individuals from the 1994 U.S. census, with sensitive attributes race and sex, as well as instances with missing values. The task is to predict if an individual earns more or less than $\$50,000$ per year.

\german -- The German Credit dataset~\footnote{\url{https://archive.ics.uci.edu/ml/support/Statlog+(German+Credit+Data)}} contains demographic and financial data about people, as well as the sensitive attribute sex. The task is to predict an individual's credit risk.

\propublica -- The ProPublica dataset~\footnote{\url{https://github.com/propublica/compas-analysis}} includes data such as criminal history, jail and prison time, demographics and COMPAS risk scores for defendants from Broward County, Florida. It includes the sensitive attributes race and sex. The prediction concerns a binary ``recidivism'' outcome, denoting whether a person was rearrested within two years after the charge given in the data. 

\ricci -- The Ricci dataset contains promotion data about firefighters, used as part of a Supreme court case (Ricci v. DeStefano) dealing with racial discrimination. The dataset contains the sensitive attribute race. The task is to predict the promotion decision.  The original promotion decision (assignment to the positive class) was made by a threshold of achieving at least a score of 70 on the combined exam outcome.

Integrating a custom dataset with \fairbench{} only requires users to load the data as a pandas dataframe and configure several class variables that denote which attributes to use as numeric and categorical features, which attribute to use as the class label, and how to identify the protected groups in the dataset.

\header{Data Preprocessing Steps} We highlight some of the data preprocessing and feature transformation operations supported by \fairbench{}. We integrate common feature scaling techniques such as standardisation and min-max scaling from \sklearn{}. We additionally provide a component that does not scale numeric features (which might be dangerous) for studying the effect of this preprocessing step. 

Additionally, we provide a \texttt{MissingValueHandler} interface to define different ways how to treat records with missing values. \fairbench{} offers a set of predefined strategies such as `complete case analysis' (removal of records with missing values) or a simple imputation strategy based on \sklearn{}'s \texttt{ModeImputer} that fills in the most frequent value of an attribute. In addition, our abstraction also supports more sophisticated techniques that learn a model to impute missing values. 
We provide an example of such a strategy as part of \fairbench{}'s code base that leverages \textit{Datawig}~\cite{Biessmann2018}, an imputation library that auto-featurizes data and learns a deep learning model tailored to the data for imputation. Its implementation focuses on imputing one column at a time for efficient modeling. We utilize this approach in the \texttt{fit} method to learn an imputation model for each feature using the remaining features (but not the class label) in the training dataset as input. At imputation time (in the \texttt{handle\_missing} method), each of of the fitted models is applied on the target data to impute the missing attributes.

\begin{python}
class DatawigImputer(MissingValueHandler)
... 
 def fit(self, train_data):
   columns = train_data.feature_columns
   # Learn an imputation model for each column
   for target_column in self.target_columns:
     input_columns = [column in columns 
                      if column != target_column]
     self.imputers[target_column] = datawig.Imputer(
       input_columns=input_columns, 
       output_column=target_col)
      .fit(train_df=train_data)

 def handle_missing(self, target_data):
   completed_data = target_data.copy()
   # Impute each column
   for column in self.target_columns:
    completed_data[column] = 
     self.imputers[column].predict(target_data)
   return completed_data      
\end{python}

\header{Models} \fairbench{} exposes a simple interface for learning algorithms, to allow the integration of many different models with low effort. The \texttt{fit\_model} method of a learner provides the implementation with access to the training data and the random seed used by the current run (to allow for reproducible training). We provide implementations for common ML models, such as logistic regression (\texttt{SGDClassifier} with logistic loss function) and decision trees from \sklearn{}. We now give two examples of integrating learners: a baseline model from \sklearn{} and an in-processing intervention from \aif{}.

\headerl{Integrating a baseline model from \sklearn{}} We implement a logistic regression learner with 5-fold cross-validation into our framework as follows. We grid search over common hyperparameters for logistic regression, such as the type of regularization and the learning rate. With the defined parameter choices and 5-fold cross validation, the grid search automatically investigates 60 different settings. Note that we propagate the random seed to all components to ensure reproducible behavior.

\begin{python}
class LogisticRegression(Learner):
 def fit_model(self, train_data, seed):
  # Hyperparameter grid
  param_grid = {
   'learner__loss': ['log'],
   'learner__penalty': ['l2', 'l1', 'elasticnet'],
   'learner__alpha': [0.00005, 0.0001, 0.005, 0.001] }
  # Pipeline for classifier
  pipe = Pipeline([
   ('learner', SGDClassifier(random_state=seed)) ])
  # Setup 5-fold cross-validation
  search = GridSearchCV(pipe, param_grid, cv=5,
    random_state=seed, 
    fit_params={'learner__sample_weight': 
               train_data.instance_weights})
  # Learn model via cross-validation               
  return search.fit(train_data.features, 
                    train_data.labels)
\end{python}

\headerl{Integrating an in-processing intervention} Next, we show how to integrate an in-processing fairness-enhancing intervention from \aif{}. Adversarial debiasing~\cite{Zhang2018} learns a classifier to maximize  prediction accuracy and simultaneously reduce an adversary's ability to determine the protected attribute from the predictions. This model can be integrated into \fairbench{}  with a few lines of code.
    
\begin{python}
class AdversarialDebiasing(Learner):
 ...
 def fit_model(self, train_data, seed):
  ad_model = AdversarialDebiasingAIF360(
   privileged_groups=self.privileged_groups,
   unprivileged_groups=self.unprivileged_groups,
   sess=self.tf_session, seed=seed)
  return ad_model.fit(annotated_train_data)
\end{python}

\begin{figure*}[t!]
     \centering
     \subfigure[Accuracy / disparate impact for {\bf \textcolor{red}{tuned}} vs. {\bf \textcolor{dgray}{non-tuned}} logistic regression models on \german.]{
       \centering
       \includegraphics[width=0.64\columnwidth]{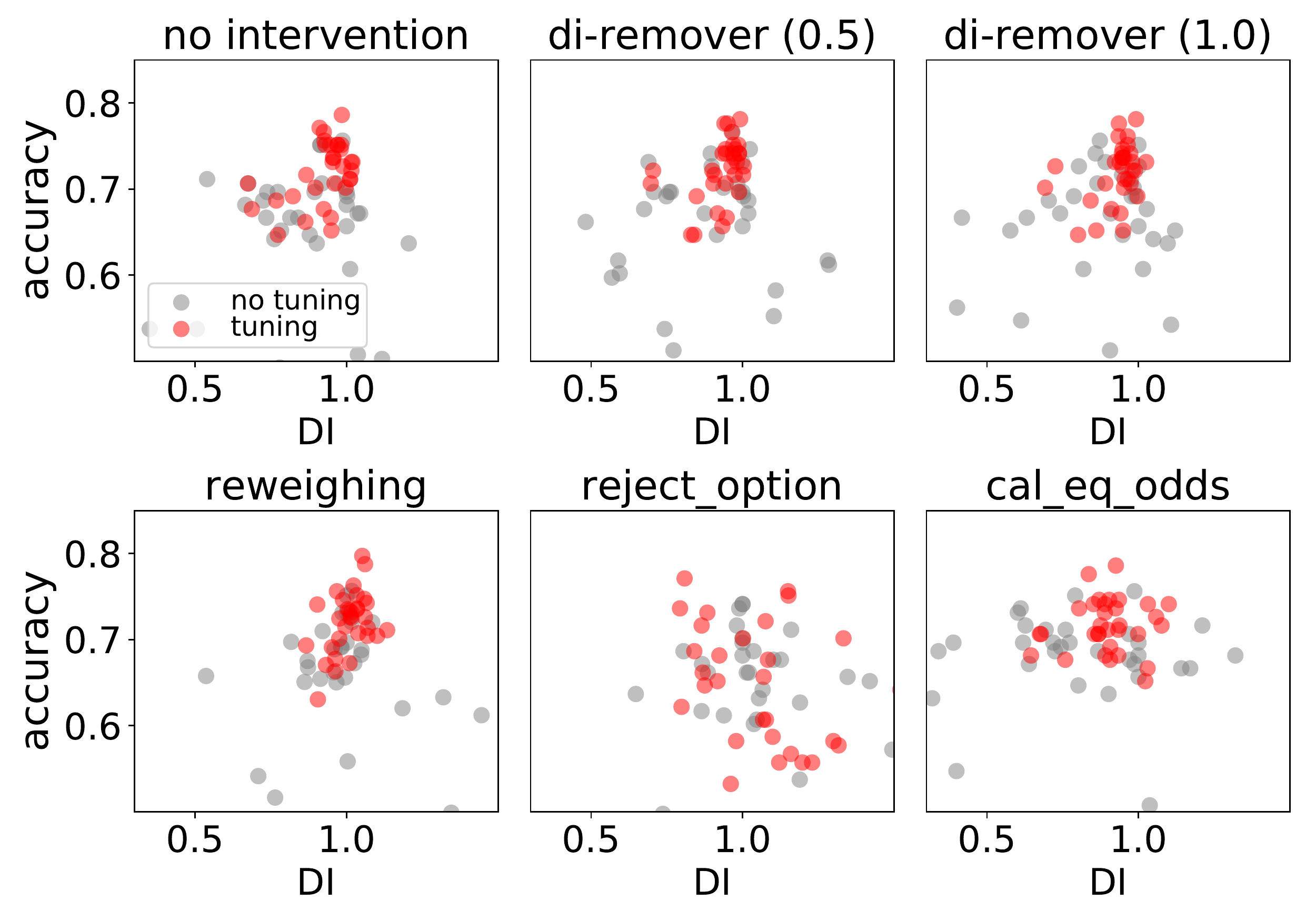}
       \label{fig:tuning-lr-di}
     } 
     \hfill
     \subfigure[Accuracy / false negative rate difference for {\bf \textcolor{red}{tuned}} vs. {\bf \textcolor{dgray}{non-tuned}} logistic regression models on \german.]{
       \centering
       \includegraphics[width=0.64\columnwidth]{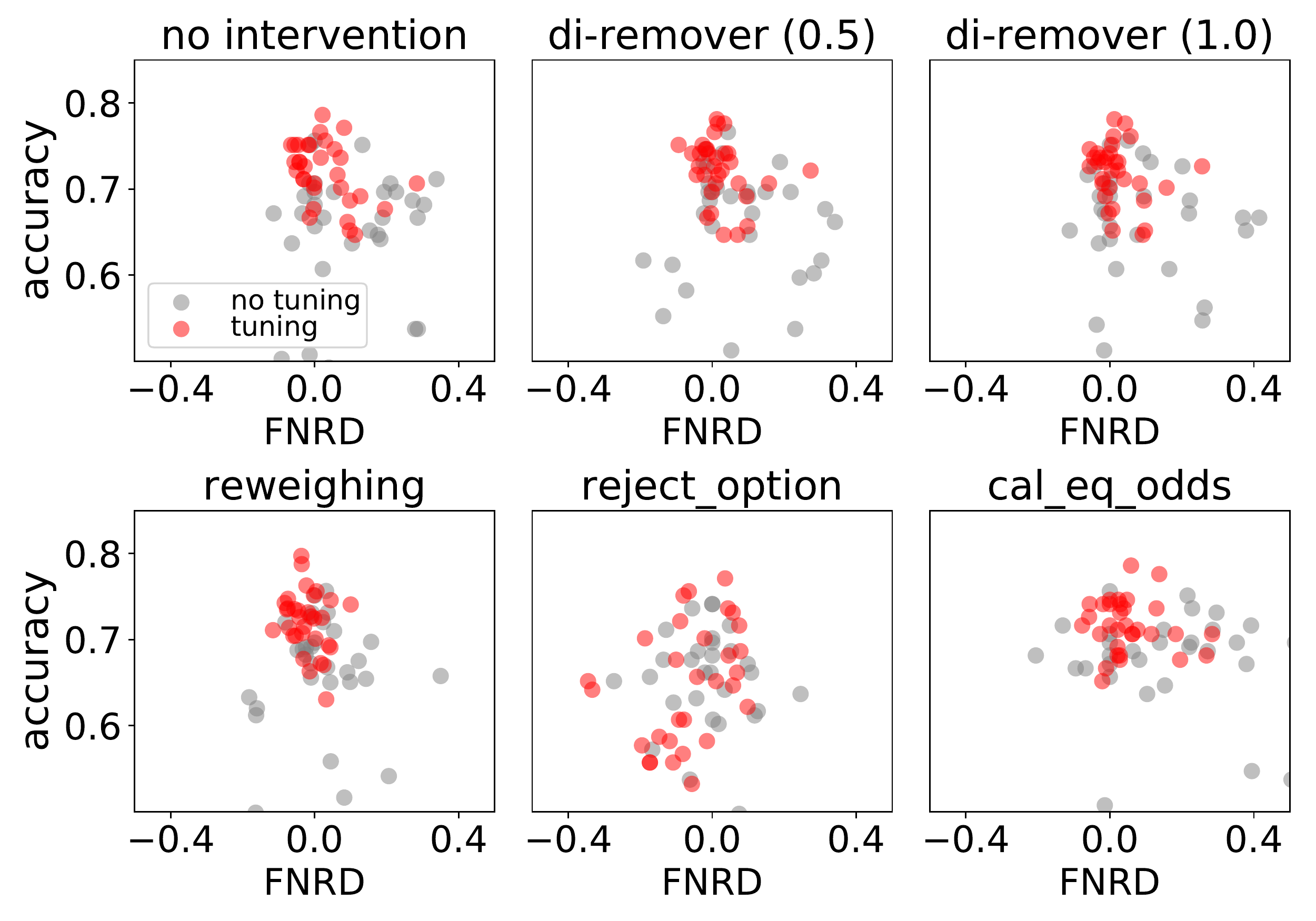}
       \label{fig:tuning-lr-fnrd}
     } 
     \hfill
     \subfigure[Accuracy / false positive rate difference for {\bf \textcolor{red}{tuned}} vs. {\bf \textcolor{dgray}{non-tuned}} logistic regression models on \german.]{
       \centering
       \includegraphics[width=0.64\columnwidth]{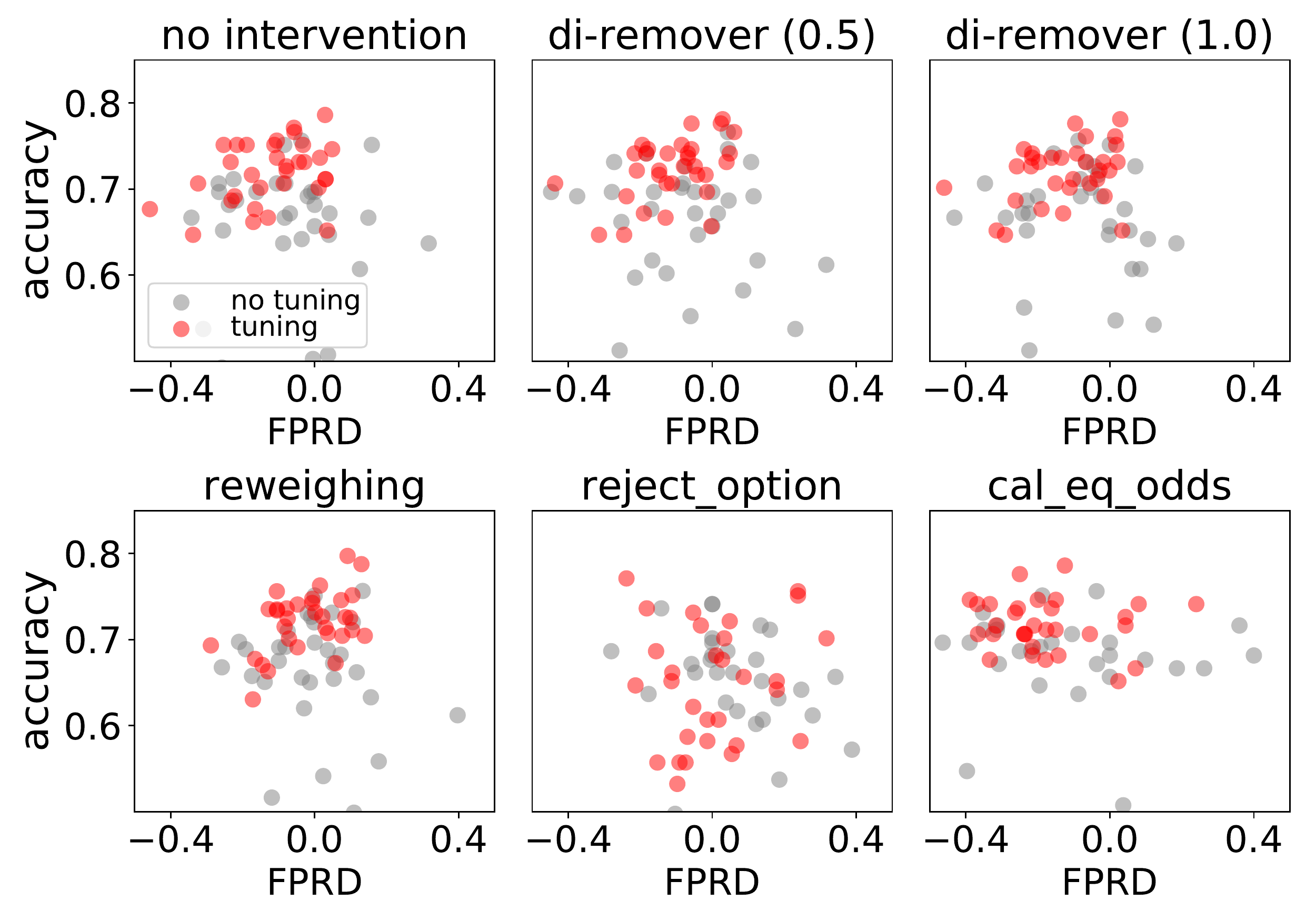}
       \label{fig:tuning-lr-fprd}
     } 
     \hfill
     \subfigure[Accuracy / disparate impact for {\bf \textcolor{red}{tuned}} vs. {\bf \textcolor{dgray}{non-tuned}} decision tree models on \german.]{
       \centering
       \includegraphics[width=0.64\columnwidth]{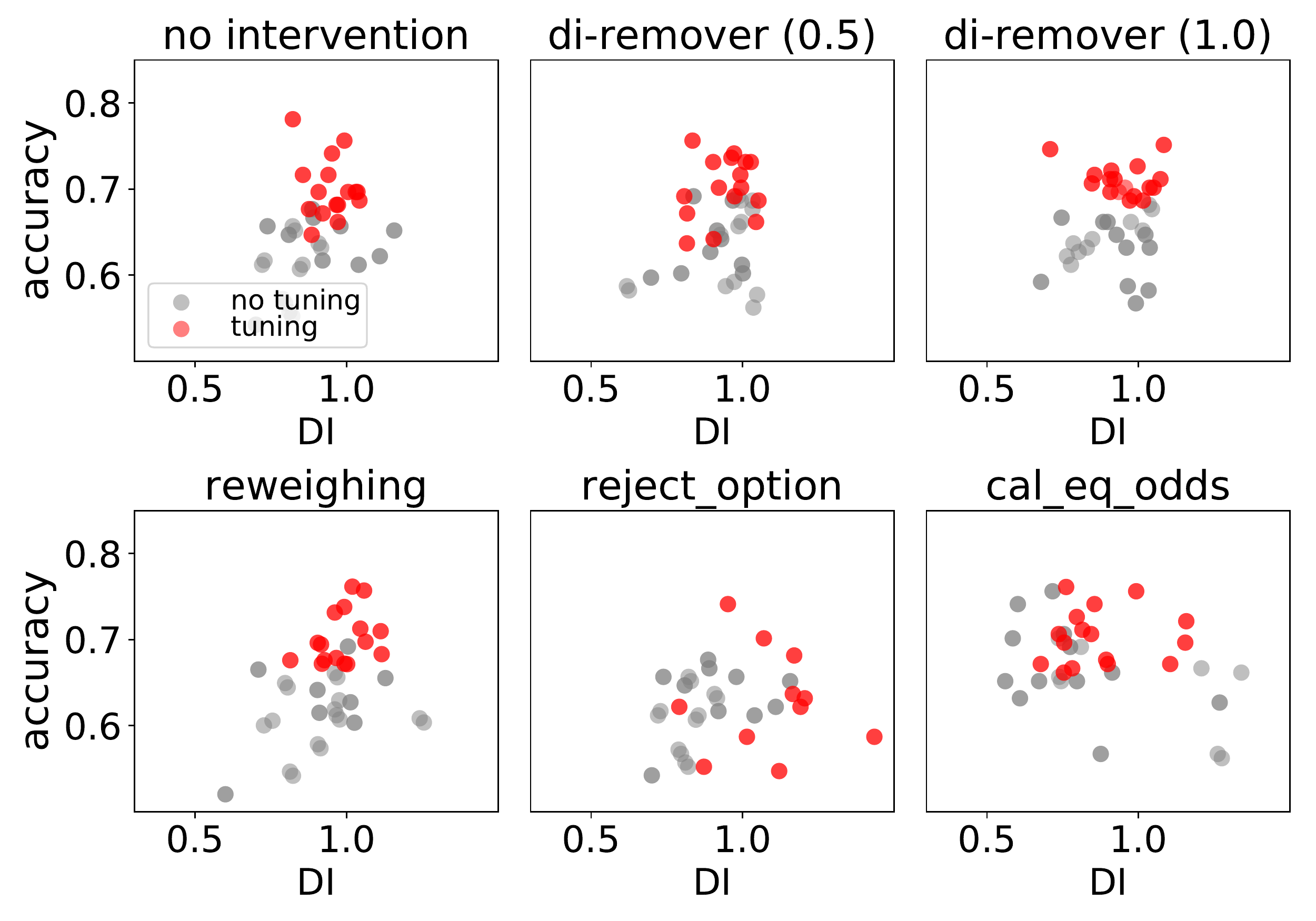}
       \label{fig:tuning-dt-di}
     } 
     \hfill
     \subfigure[Accuracy / false negative rate difference for {\bf \textcolor{red}{tuned}} vs. {\bf \textcolor{dgray}{non-tuned}} decision tree models on \german.]{
       \centering
       \includegraphics[width=0.64\columnwidth]{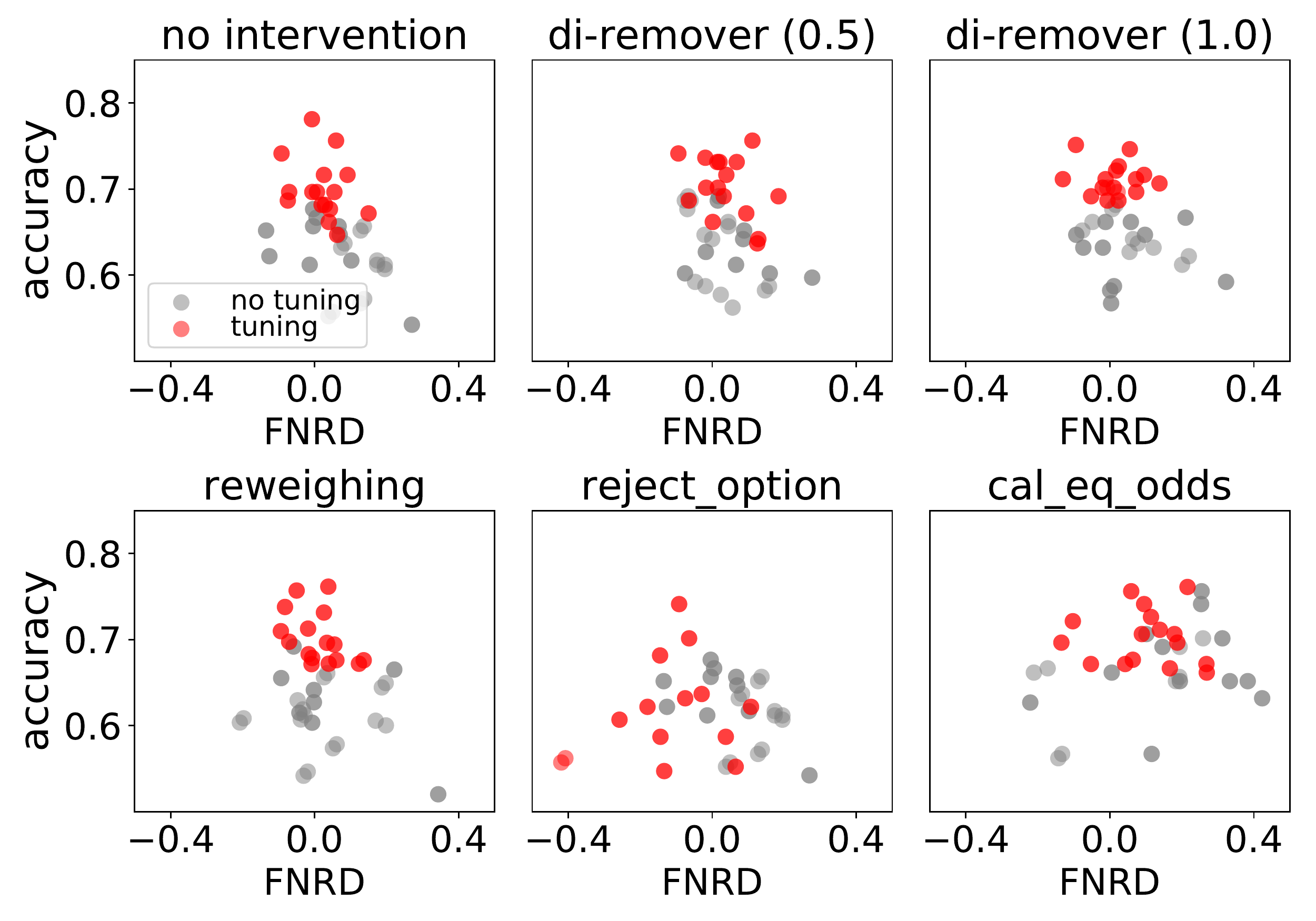}
       \label{fig:tuning-dt-fnrd}
     } 
     \hfill
     \subfigure[Accuracy / false positive rate difference for {\bf \textcolor{red}{tuned}} vs. {\bf \textcolor{dgray}{non-tuned}} decision tree models on \german.]{
       \centering
       \includegraphics[width=0.64\columnwidth]{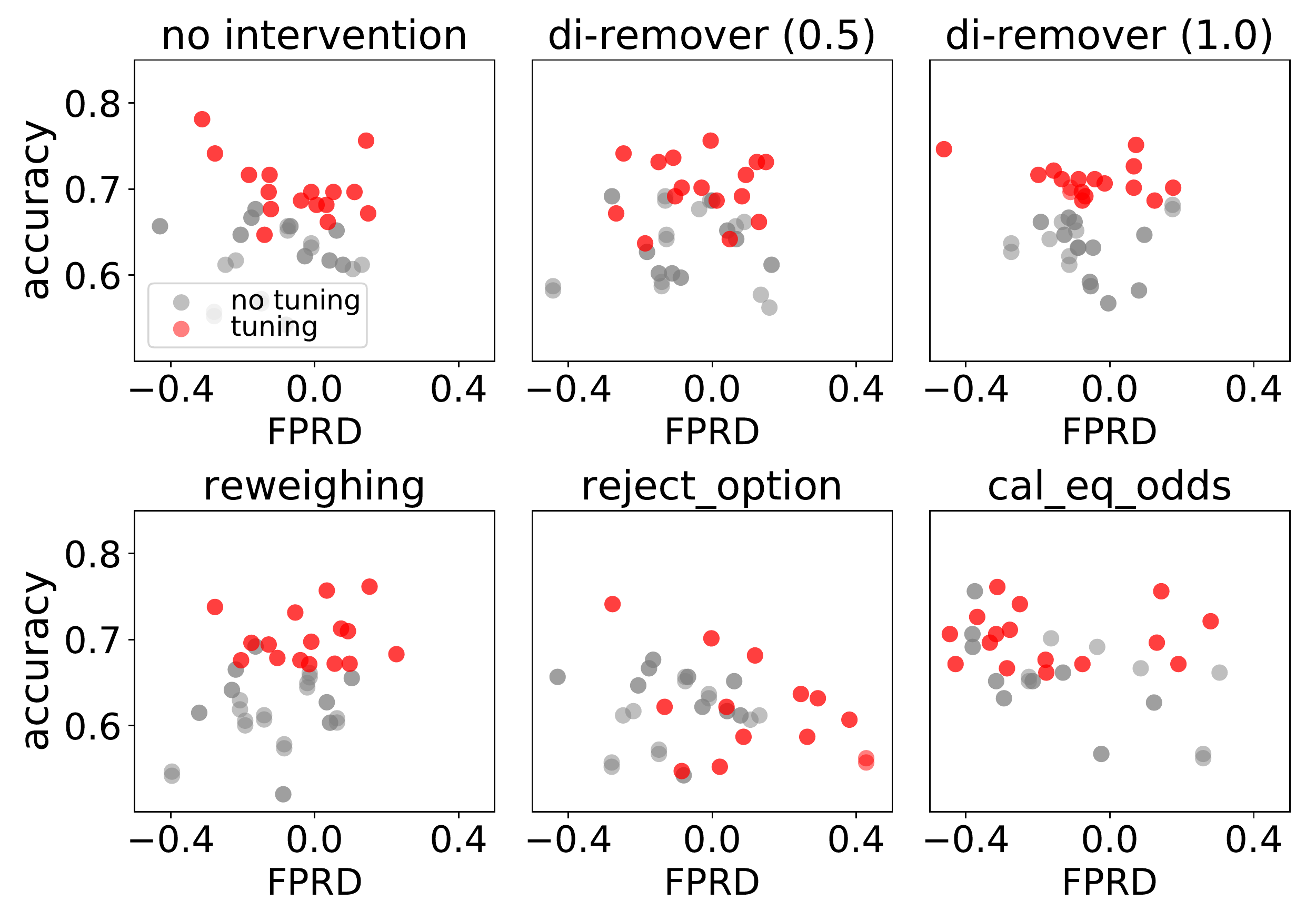}
       \label{fig:tuning-dt-fprd}
     }           
     \hfill       
     \caption{Impact of hyperparameter tuning on the accuracy and fairness metrics of logistic regression and decision tree models (in combination with various preprocessing and postprocessing interventions) on the \texttt{germancredit} dataset. Hyperparameter tuning (\textcolor{red}{red dots}) results in higher accuracy and reduced variance of the fairness outcome compared to no tuning (\textcolor{dgray}{gray dots}) in many cases.}        
     \label{fig:tuning}
\end{figure*}

\header{Fairness Enhancing Interventions} Next, we focus on fairness-enhancing interventions. Note that in-processing methods, which learn a specialized model, can simply be implemented as learners into our framework. Therefore, we only need to additionally integrate the pre-processing and post-processing interventions, for which \fairbench{} provides dedicated abstractions. 
In the following, we detail how to integrate a preprocessing technique called `disparate impact removal'~\cite{DBLP:conf/kdd/FeldmanFMSV15}, which edits feature values to increase group fairness while preserving the rank-ordering within groups. The repair level parameter represents the repair amount. We leverage the implementation from \aif{}. Note that \fairbench{} provides information about protected and unprotected groups in the dataset to the preprocessing intervention. The integration of post-processing techniques works analogously.
\begin{python}
class DIRemover(Preprocessor):
  def pre_process(self, data, privileged_groups, 
                  unprivileged_groups, seed):
   diremover = DisparateImpactRemover(
     repair_level=self.repair_level)
   return diremover.fit_transform(annotated_data)
\end{python}

\header{Metrics} We leverage the metrics implementations from \aif{}\footnote{\url{https://github.com/IBM/AIF360/tree/master/aif360/metrics}}, and compute 25 different metrics for the overall train and test set, as well as separately for the privileged and unprivileged groups. In adddition, we compute 22 different global metrics that measure the effects between the privileged and the unprivileged groups. Every experiment writes an output file with these metrics by default.

\header{Example} We finally revisit our introductory example from Section~\ref{sec:intro:example}, where our data scientist Ann wants to investigate the impact of different fairness-enhancing interventions on her classifier that decides on payment option offerings. 

\begin{python}
# Fixed random seeds for reproducibility
seeds = [46947, 71735, 94246, ...]
# Interventions
interventions = [NoIntervention(), Reweighing(), 
                 DiRemover(0.5)]
for seed in seeds:
 for intervention in interventions:
  # Configure experiment
  exp = PaymentOptionGenderExperiment(
   random_seed=seed, 
   missing_value_handler=DatawigImputer('age'),
   numeric_attribute_scaler=StandardScaler(),
   learner=LogisticRegression(), 
   pre_processor=intervention)
  # run experiment, and write metrics to disk  
  exp.run()
\end{python}

\noindent Ann integrates her custom dataset via a \texttt{PaymentOptionExperiment} class with \fairbench{}, which describes how to load the dataset and defines the attributes to use as features, as the label, and the sensitive attributes. Next, she configures the experiment to match her use case by defining a logistic regression model as the baseline algorithm, leveraging Datawig to impute the \texttt{age} attribute of customers, and appropriately scaling features.
Furthermore, she defines a set of pre-processing interventions, for which she would like to investigate the impact on the outcome of the classifier, and fixes the set of random seeds to use, to have reproducible results. Now she is ready to run the experiments, which will output the resulting metrics to disk (so that they can be subsequently explored via a juypter notebook).
\section{Experimental Evaluation}
\label{sec:study}

We now demonstrate how \fairbench{} can be used to showcase and overcome some of the shortcomings outlined in Section~\ref{sec:shortcomings}.

\subsection{Impact of Hyperparameter Tuning on the Variability of Accuracy and Fairness}

In the first experiment, we aim to investigate the effect of not tuning the hyperparameters of baseline models during experimentation (as discussed in Section~\ref{sec:shortcomings:tune}).

\header{Dataset} We leverage the \german dataset for this experiment, which contains 20 demographic and financial attributes of 1000 people, as well as the sensitive attribute sex. The task is to predict each individual's credit risk.

\header{Setup} We configure \fairbench{} as follows: we randomly split the data into 70\% train data, 10\% validation data, 20\% test data, based on supplied fixed random seeds (for reproducibility).  We apply a fixed set of data preprocessing steps: we do not resample the data, do not handle missing values (as the data is complete already), and standardize numeric features. We leverage two baseline models (logistic regression and decision trees) in two different variants each: $(i)$ without hyperparameter tuning, where we just use the default hyperparameters of the baseline model; $(ii)$ with hyperparameter tuning, where apply gridsearch (over 3 regularizers and 4 learning rates for logistic regression; over 2 split criteria, 3 depth params, 4 min samples per leaf params, 3 min samples per split params for the decision tree) and five-fold cross validation on the training data. We apply three different fairness-enhancing interventions that preprocess the data: `disparate impact remover' (`di-remover' in the plots)~\cite{DBLP:conf/kdd/FeldmanFMSV15} with repair levels 0.5 and 1.0, as well as `reweighing'~\cite{Kamiran2012}. Additionally, we experiment with two different fairness-enhancing interventions that post-process the predictions: `reject option classification'~\cite{Kamiran2012b} and `calibrated equal odds'~\cite{Pleiss2017}. We leverage 16 different random seeds for the experiment and execute 1,344 runs in total. We report metrics computed from the predictions on the held-out test set.

\header{Results} We plot the results of this experiment in Figure ~\ref{fig:tuning}, where we show the resulting accuracy and several fairness related measures\footnote{Note that we plot these measures regardless of whether the intervention optimizes for them or not.} between the privileged and unprivileged groups, including disparate impact (DI), the difference in false negative rates (FNRD), and the difference in false positive rates (FPRD). The red dots denote the outcome when we apply hyperparameter tuning to the baseline model, while the gray dots denote the outcome using the default model parameters, without tuning.

We observe a large number of cases where the tuned variant results in both, a higher accuracy model and a lower variance in the fairness outcome. Examples are $(i)$~the accuracy and disparate impact for the `di-remover' and `reweighing' interventions of both logistic regression and decision trees in Figures~\ref{fig:tuning-lr-di}~\&~\ref{fig:tuning-dt-di}, $(ii)$~the accuracy and false negative rate difference for `di-remover' and logistic regression, `di-remover' and decision trees, as well as 'reweighing' and decision trees in Figures~\ref{fig:tuning-lr-fnrd}~\&~\ref{fig:tuning-dt-fnrd}; and $(iii)$~accuracy and false positive rate difference for `di-remover' for both models and `reweighing' for the decision tree.

These results strongly suggest that the high variability of the fairness and accuracy outcomes with respect to different train/test splits observed in previous studies~\cite{DBLP:conf/fat/FriedlerSVCHR19} might be an artifact of the lack of hyperparameter tuning of the baseline models in these studies~(as discussed in Section~\ref{sec:shortcomings:tune}).

\begin{figure}[t!]
     \centering
     \subfigure[\small{High sensitivity on feature scaling for logistic regression on the \ricci dataset.  {\bf \textcolor{red}{Red dots}} represent experiments with feature scaling, {\bf \textcolor{dgray}{gray dots}} represent experiments without scaling, and contain many cases where no satisfactory classifier could be learned.}]{
       \centering
       \includegraphics[width=\columnwidth]{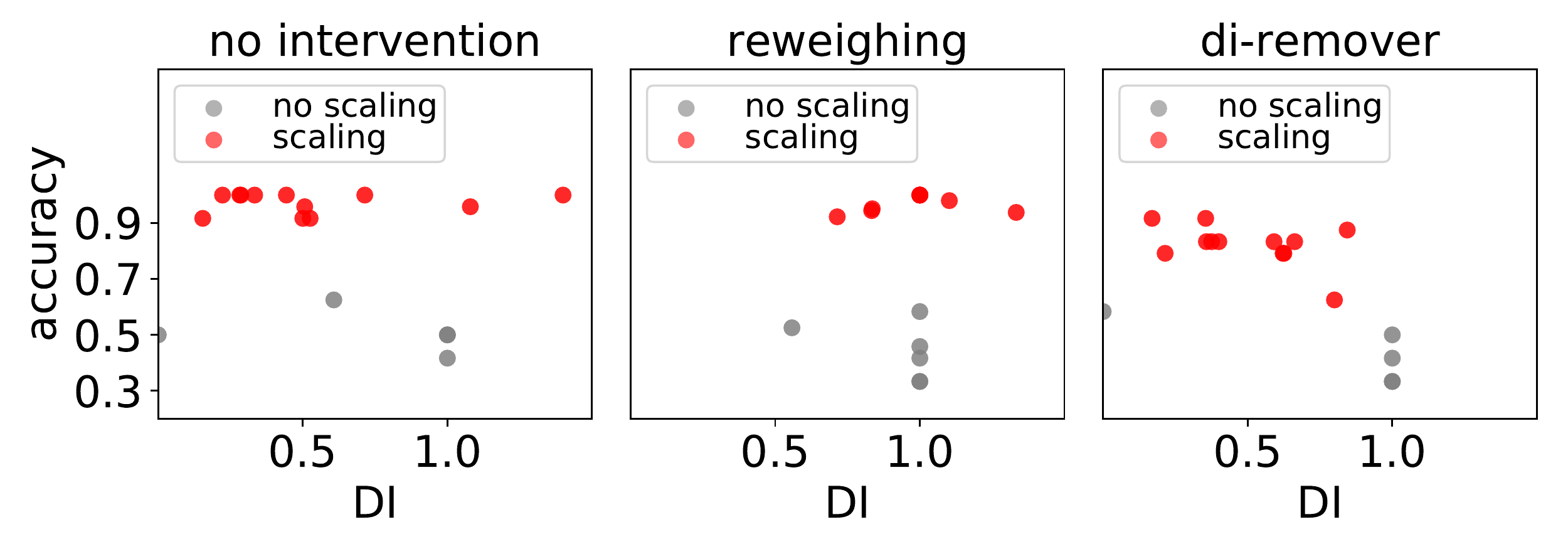}
       \label{fig:scaling-lr}
     } 
     \hfill
     \subfigure[\small{Robustness of decision trees against a lack of feature scaling on the \ricci dataset.  {\bf \textcolor{red}{Red dots}} represent experiments with feature scaling, {\bf \textcolor{dgray}{gray dots}} represent experiments without feature scaling.}]{
       \centering
       \includegraphics[width=\columnwidth]{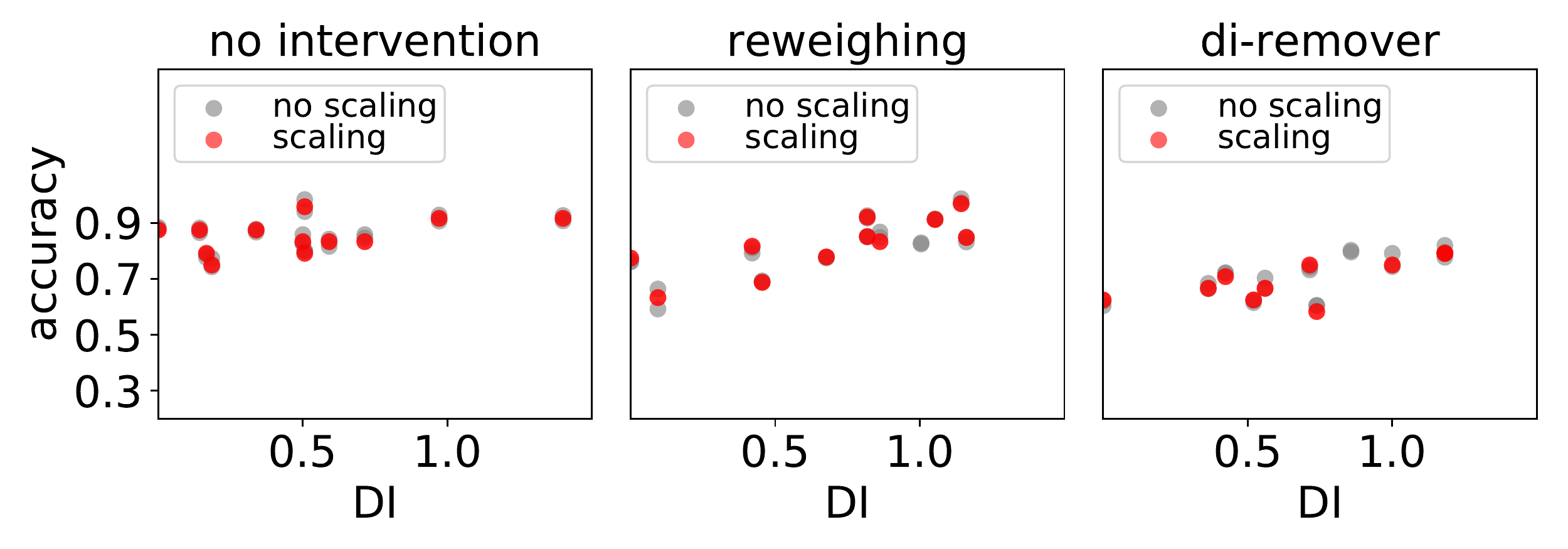}
       \label{fig:scaling-dt}
     } 
     \caption{Comparison of the impact of a lack of numeric feature scaling on the \ricci dataset. Decision trees are robust against this lack, while logistic regression often fails to learn a reasonable classifier.}        
     \label{fig:scaling}
\end{figure}

\subsection{Impact of Feature Scaling}

In the next experiment, we show that the lack of feature scaling~(Section~\ref{sec:shortcomings:scale}) can lead to the failure to learn a well-working model.

\header{Dataset} We leverage the \ricci dataset that has 118 entries and five attributes, including the sensitive attribute race.

\header{Setup} We configure \fairbench{} as follows: we randomly split the data into 70\% train data, 10\% validation data, 20\% test data, based on supplied random seeds. We do not resample the data, and do not need to handle missing values as the data is complete already. We leverage two baseline models (logistic regression and decision tree), with hyperparameter tuning analogous to the previous section, and two fairness-enhancing interventions that preprocess the data: `disparate impact remover' and `reweighing'. We vary the treatment of numeric features, however: for one set of runs, we leave these on their original scale, and for the remaining runs we standardize them using our integration of \sklearn{}'s \texttt{StandardScaler}. We execute 216 runs in total and report metrics from predictions on the held-out test set.

\header{Results} The results of our runs are illustrated in Figure~\ref{fig:scaling}, where Figure~\ref{fig:scaling-lr} shows the accuracy and disparate impact for the logistic regression model (under different interventions) and Figure~\ref{fig:scaling-dt} analogously shows the results for the decision tree. The lack of scaling did not impact the results for the decision tree: the red dots with feature scaling and the gray dots without feature scaling are overlapping.  Logistic regression (trained with stochastic gradient descent in this setup), however, often fails to learn a valid model in the case of unscaled features, resulting in an accuracy under 50\% --- worse than if classification decisions were assigned at random. This finding confirms our claim that a lack of feature scaling alone can lead to unsatisfactory results in an ML evaluation setup (independently of fairness-related issues).

\begin{figure}[t!]
     \centering
     \subfigure[Accuracy for different missing value imputation strategies and a logistic regression baseline model on the \adult dataset. Imputed records are denoted by {\bf \textcolor{red}{red dots}}, complete records are denoted by {\bf \textcolor{dgray}{gray dots}}. No significant difference between mode and datawig imputation is observed.]{
       \centering
       \includegraphics[width=\columnwidth]{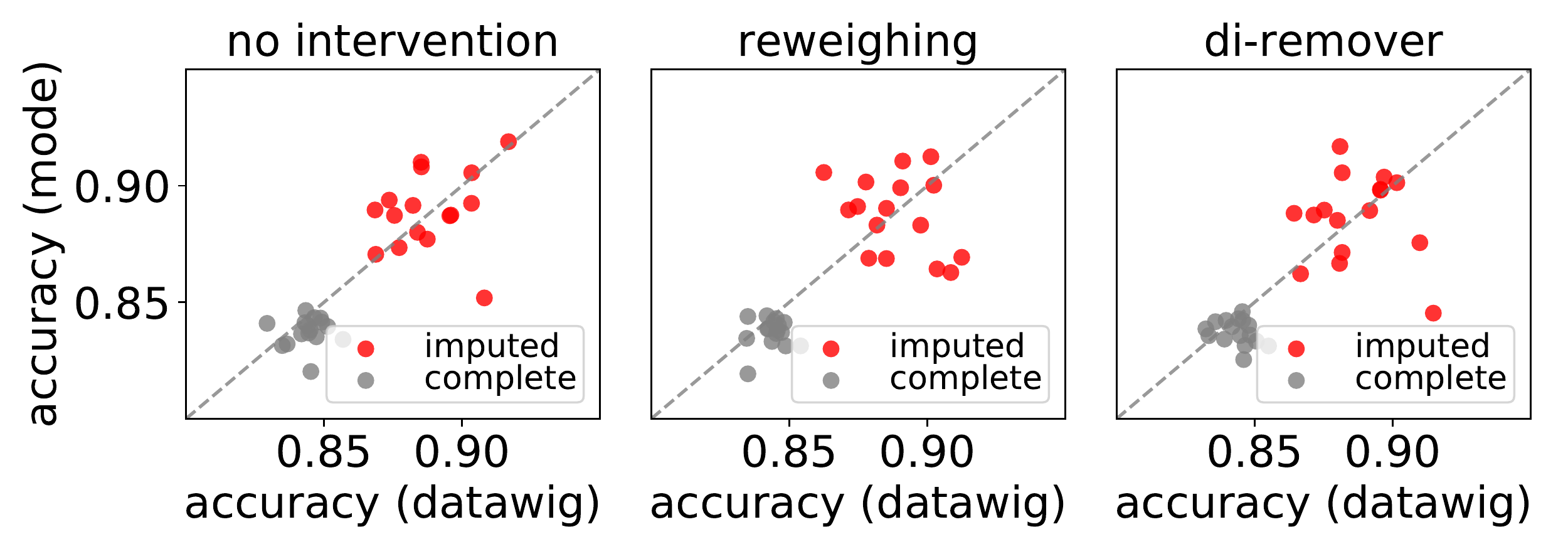}
       \label{fig:missing-acc-separate-dt}
     }                 
     \hfill
     \subfigure[Accuracy for different missing value imputation strategies and a decision tree baseline model on the \adult dataset. Imputed records are denoted by {\bf \textcolor{red}{red dots}}, complete records are denoted by {\bf \textcolor{dgray}{gray dots}}.  No significant difference between mode and datawig imputation is observed.]{
       \centering
       \includegraphics[width=\columnwidth]{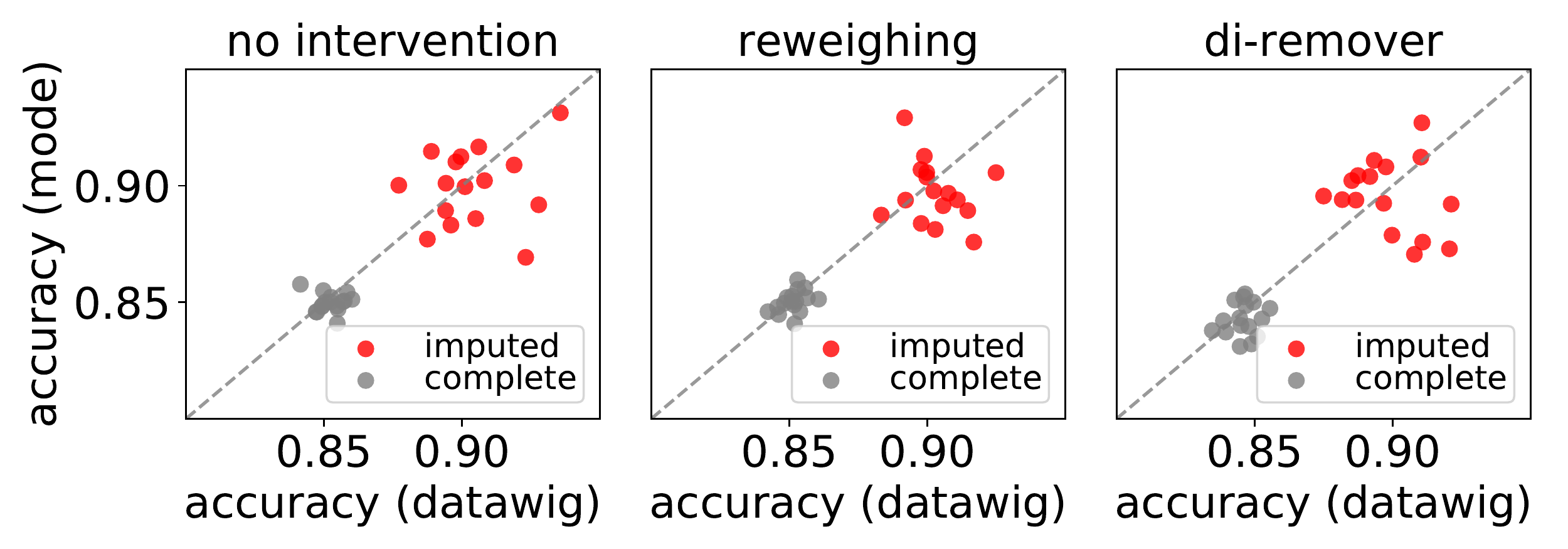}
       \label{fig:missing-acc-separate-lr}
     } 
     \caption{Impact of missing value imputation on the prediction accuracy for different imputation strategies and interventions. We observe a higher accuracy for incomplete records ({\bf \textcolor{red}{red dots}}), which we attribute to the fact that data is not missing and random, and incomplete records contain more easy-to-classify negative examples.}        
     \label{fig:missing-acc-separate}
\end{figure}

\subsection{Impact of Missing Value Imputation}

In our third experiment, we showcase how \fairbench{} can be leveraged to investigate the effects of including records with missing values into a study (which are commonly filtered out in other studies and toolkits, as discussed in Section~\ref{sec:shortcomings:missing}).

\header{Dataset} We leverage the \adult dataset for this experiment, with 32,561 instances and 14 attributes, including the sensitive attributes \texttt{race} and \texttt{sex}, and 2,399 instances with missing values. The task is to predict whether an individual makes more or less than $\$50,000$ per year. Fairness evaluation is conducted between the privileged group of white individuals (85\% of records) and the unprivileged group of non-white individuals (15\% of records).

Among the 14 attributes, three have missing values --- \texttt{workclass}, \texttt{occupation}, and \texttt{native-country}. Missing values do not seem to occur at random, as the records with missing values exhibit very different statistics than the complete records. For example, the positive class label (high income) occurs with 24\% probability among the complete records, but only with 14\% probability in the records with missing values. Additionally, married individuals are in the vast majority in the complete records, while the most frequent \texttt{marital-status} among the incomplete records is \textit{never-married}.

Furthermore, the records with missing values from the privileged group are very different from the records with missing values from the unprivileged group. For example, the attribute \texttt{native-country} is missing  four times more frequently for non-white individuals than for white individuals. Among the incomplete privileged records there is a 15\% chance of a high income, the second largest age group consists of 60 to 70 year-olds, and the majority of the individuals is married. For the incomplete records from the non-privileged group however, there is only a 10.6\% chance of a high income, it contains very few seniors, and the majority of the individuals is unmarried.

\begin{figure}[t!]
     \centering
     \subfigure[Impact of the inclusion of imputed records ({\bf \textcolor{red}{red dots}}) on the accuracy and disparate impact of a logistic regression model and various interventions on \adult.]{
       \centering
       \includegraphics[width=\columnwidth]{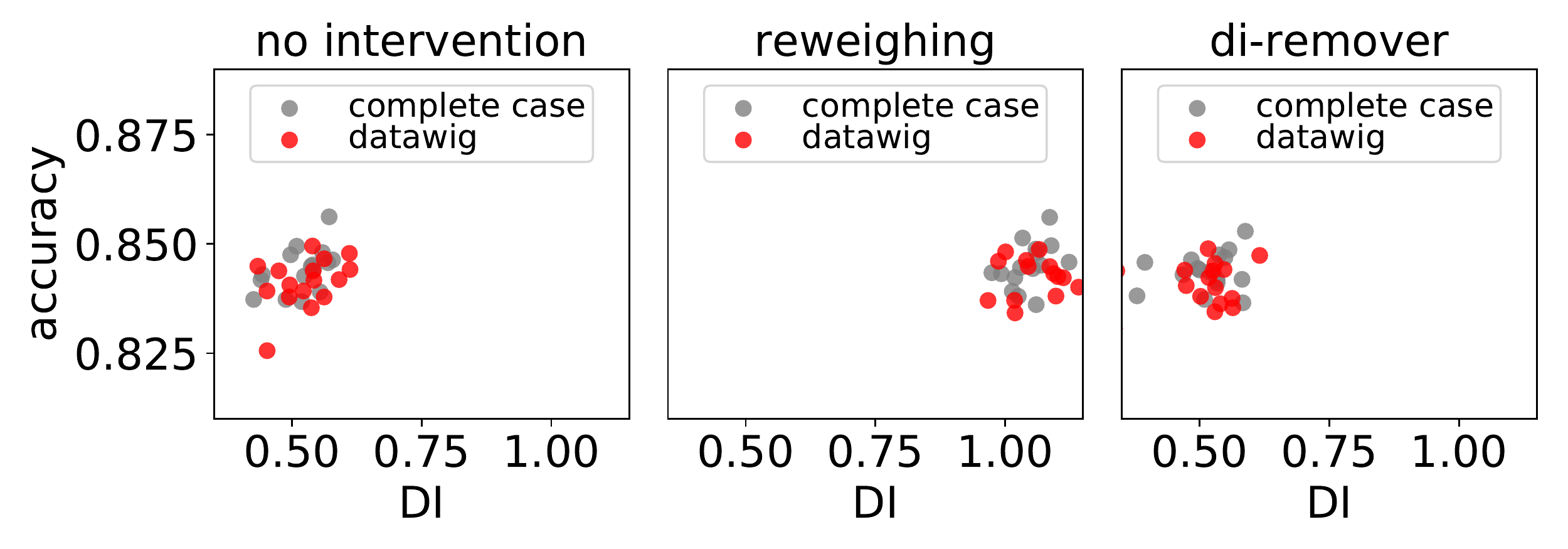}
       \label{fig:missing-acc-di-dt}
     }                 
     \hfill
     \subfigure[Impact of the inclusion of imputed records ({\bf \textcolor{red}{red dots}}) on the accuracy and disparate impact of a decision tree model and various interventions on \adult.]{
       \centering
       \includegraphics[width=\columnwidth]{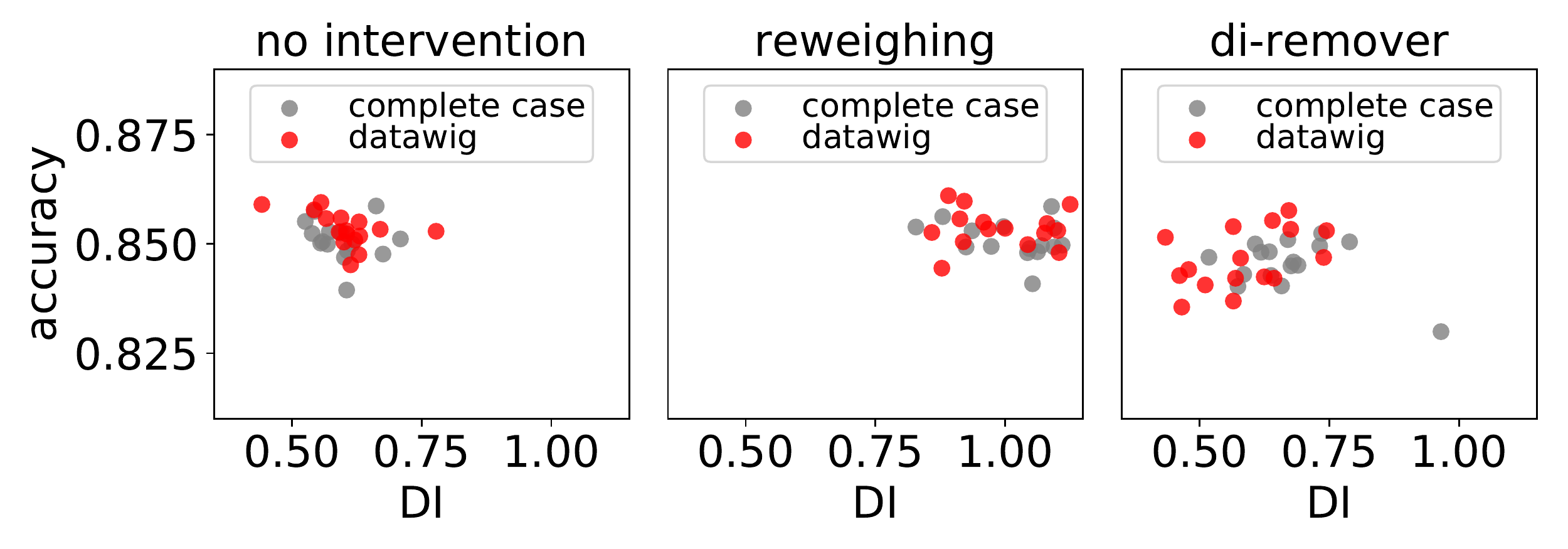}
       \label{fig:missing-acc-di-lr}
     } 
     \caption{Comparison of accuracy and disparate impact on the \adult dataset for complete case analysis (removal of incomplete records, {\bf \textcolor{dgray}{gray dots}}) and inclusion of incomplete records (with imputation of missing values via datawig, {\bf \textcolor{red}{red dots}}). Including imputed records does not significantly affect the disparate impact of the resulting models.}        
     \label{fig:missing-acc-di}
\end{figure}

\header{Setup} We configure \fairbench{} as follows: we randomly split the data into 70\% train data, 10\% validation data, 20\% test data, based on supplied random seeds. We do not resample the data, standardize numerical features and leverage logistic regression and decision trees as baseline learners with hyperparameter tuning analogous to previous experiments. We apply two different fairness enhancing interventions that preprocess the data: `disparate impact remover' and `reweighing'. We vary the strategy to treat missing values for this experiment: $(i)$ we apply complete case analysis and remove incomplete records; $(ii)$ we retain all records and impute missing values with `mode imputation'; $(iii)$ we retain all records and apply model-based imputation with datawig~\cite{Biessmann2018}. We execute 530 runs in total, and again report metrics from predictions on the held-out test set.

\header{Results} Figure~\ref{fig:missing-acc-separate} shows classification accuracy for complete (gray dots) and incomplete (red dots) records, under imputation with mode and datawig. First, we observe that records with imputed values achieve high accuracy.  This is a significant result, since these records could not have been classified at all before imputation!  Interestingly, we observe higher accuracy for records with missing values compared to the complete records.  Based on our understanding of the data, described earlier in this section, we attribute this to the higher fraction of (easier to classify) negative examples among the incomplete records. Further, we do not observe a significant difference in accuracy between mode imputation and datawig imputation. We attribute this to the highly skewed distribution of the attributes to impute ---  a favorable setting for mode imputation. Because datawig does no worse than mode, and is expected to perform better in general~\cite{Biessmann2018}, we only present results for datawig-based imputation in the next, and final, plot.

Figure~\ref{fig:missing-acc-di} shows the accuracy and disparate impact of complete case analysis (\eg  the removal of incomplete records) versus the inclusion of incomplete records with datawig imputation. We observe a minimally higher accuracy in the case of including incomplete records, but in general find no significant positive or negative impact on disparate impact.  Taken together, the results in Figures~\ref{fig:missing-acc-separate} and~\ref{fig:missing-acc-di} paint an encouraging picture: Imputation allows us to classify records with missing values, and do so accurately, and it does not degrade performance, either in terms of accuracy or in terms of fairness, for the complete records.

\section{Related Work}
\label{sec:related}

As the research on algorithmic fairness begins to mature, efforts are being made to standardize the requirements~\cite{DBLP:conf/icse/KoeneDS18}, systematize the measures and the algorithms~\cite{21defs,DBLP:journals/datamine/Zliobaite17},  and generalize the insights~\cite{DBLP:journals/corr/Chouldechova17,DBLP:journals/corr/FriedlerSV16,DBLP:conf/sigmetrics/Kleinberg18}.  As we are preparing to translate the research advances made by this community into data science practice, it is essential that we develop methods for judicious evaluation of our techniques, and for integrating them into real-world testing and deployment scenarios.  Our work on \fairbench{} is motivated by this need. 

{\bf The importance of benchmarking.} Other systems communities have benefited tremendously from benchmarking and standardisation efforts.  For example, the investment of the National Institute of Standards and Technology (NIST) into Text REtrieval Conference (TREC) started in 1992 and brought tremendous benefits both to the Information Retrieval community, and to the global economy~\cite{trec_impact}.  The Transaction Processing Performance Council (TPC), established in 1988 to develop transaction processing and database benchmarks, has had similar impact on the wide-spread commercial adoption of relational data management technology~\cite{tpc}.  While the algorithmic fairness community may not yet be ready for standardization --- considering that our understanding of the fairness measures and their trade-offs is still evolving --- we are certainly ready to grow up past the initial stage of wild exploration, and into developing and adhering to rigorous evaluation and software engineering best practices.

{\bf Design and evaluation frameworks for fairness.} In our work on \fairbench{} we build on the efforts of Friedler et al.~\cite{DBLP:conf/fat/FriedlerSVCHR19} to develop a generalizable methodology for comparing performance of fairness-enhancing interventions, and on the work of Bellamy et al.~\cite{Bellamy2018} to provide a standardized implementation framework for these methods.  We are also inspired by Stoyanovich et al.~\cite{DBLP:conf/ssdbm/StoyanovichHAMS17}, who advocate for systems-level support for responsibility properties through the data lifecycle.   
Other relevant efforts include FairTest~\cite{DBLP:conf/eurosp/TramerAGHHHJL17}, a methodology and a framework for identifying  ``unwarranted associations'' that may correspond to unfair, discriminatory, or offensive user treatment in data-driven applications. FairTest automatically discovers associations between outcomes and sensitive attributes, and provides debugging capabilities that let programmers rule out potential confounders for observed unfair effects.
The fairness-aware programming project~\cite{DBLP:conf/fat/AlbarghouthiV19} also shares motivation with our work, in that it develops a methodology for handling fairness as a systems requirement. Specifically, the authors develop a specification language that allows programmers to state
fairness expectations natively in their code, and have a runtime system monitor decision-making and report violations of fairness.  These statements are then translated to Python decorators, wrapping and modifying function behavior.  The main difference with our approach is that we do not assume a homogeneous programming environment, but rather incorporate fairness interventions into data-rich machine learning pipelines, while paying close attention to data pre-processing. Another relevant line of work is Themis~\cite{DBLP:conf/sigsoft/GalhotraBM17}, a software testing framework that automatically designs fairness tests for black-box systems.

\header{General challenges in end-to-end machine learning} Software systems that learn from data using machine learning (ML) are being deployed in increasing numbers in the real world. The operation and monitoring of such systems introduces novel challenges, which are very different from the challenges encountered in traditional data processing systems~\cite{Sculley2015,Kumar2016}. ML systems in the real world exhibit a much higher complexity than ``text book'' ML scenarios (\eg training a classifier on a standard benchmark dataset). Real world systems not only have to learn a single model, but must define and execute a whole ML pipeline, which includes data preprocessing operations such as data cleaning, standardisation and feature extraction in addition to learning the model, as well as methods for hyperparameter selection and model evaluation. Such ML pipelines are typically deployed in systems for \textit{end-to-end machine learning}~\cite{Baylor2017,Schelter2018c}, which require the integration and validation of raw input data from various input sources, as well as infrastructure for deploying and serving the trained models. These systems must also manage the lifecycle of data and models in such scenarios~\cite{Polyzotis2018}, as new (and potentially changing) input data has to be continuously processed, and the corresponding ML models have to be retrained and managed accordingly.

Many of the challenges incurred by end-to-end ML are only recently attracting the attention of the academic community.  These include enabling industry practitioners to improve the fairness in real world ML systems~\cite{Holstein2019,DBLP:conf/fat/AlbarghouthiV19}, efficiently testing and debuging ML models~\cite{Pei2017,Chung2019}, and recording the metadata and the lineage of ML experiments~\cite{Vanschoren2014,Vartak2016,Schelter2017}.  We contribute to this line of work by presenting a framework that brings the insights from end-to-end machine learning to the fairness, accountability, and transparency community.
\section{Conclusions \& Future Work}
\label{sec:conc}

We identified shortcomings in existing empirical studies and toolkits for analyzing fairness-enhancing interventions. Subsequently, we presented the design and implementation of our evaluation framework \fairbench{}. This framework empowers data scientists and software developers to configure and customise experiments on fairness-enhancing interventions with low effort, and enforces best practices in software engineering and machine learning at the same time.
We demonstrated how \fairbench{} can be leveraged to measure the impact of sound best practices, such as hyperparameter tuning and feature scaling, on the fairness and accuracy of the resulting classifiers. Additionally, we showcased how \fairbench{} enables the inclusion of incomplete data into studies (through data cleaning methods such as missing value imputation), and helps to analyze the resulting effects.

\header{Future work} We aim to extend \fairbench{} by integrating additional fairness-enhancing interventions~\cite{Salimi2019,Huang2019},  datasets,  preprocessing techniques (such as stratified sampling), and  feature transformations (such as embeddings of the input data). Additionally, we intend to extend its scope to scenarios beyond binary classification.
Furthermore, we would like to strengthen the \textit{human-in-the-loop} character of \fairbench{} by adding visualisations and allowing end-users to control  experiments with low effort. While our current focus is on data scientists and software developers as end-users, we think that it is also crucial to empower less technical users to conduct fairness-related studies~\cite{LehrOhm2017}.

\balance

\bibliographystyle{ACM-Reference-Format}
\bibliography{fp}

\end{document}